\pdfoutput=1

\documentclass[11pt]{article}
\usepackage{colortbl}
\usepackage[]{emnlp2021}

\usepackage{times}
\usepackage{latexsym}
\usepackage{graphicx,paralist,multirow}
\usepackage{amsmath}

\usepackage[T1]{fontenc}
\usepackage[utf8]{inputenc}

\usepackage{microtype}

\title{Zero-Shot Dialogue Disentanglement \\by \\ 
Self-Supervised Entangled Response Selection}

\author{Ta-Chung Chi \\
  Language Technologies Institute \\
  Carnegie Mellon University \\
  \texttt{tachungc@andrew.cmu.edu} \\\And
  Alexander I. Rudnicky \\
  Language Technologies Institute \\
  Carnegie Mellon University \\
  \texttt{air@cs.cmu.edu} \\}

\begin{document}
\maketitle
\begin{abstract}
Dialogue disentanglement aims to group utterances in a long and multi-participant dialogue into threads. This is useful for discourse analysis and downstream applications such as dialogue response selection, where it can be the first step to construct a clean context/response set.
Unfortunately, labeling all~\emph{reply-to} links takes quadratic effort w.r.t the number of utterances: an annotator must check all preceding utterances to identify the one to which the current utterance is a reply.
In this paper, we are the first to propose a~\textbf{zero-shot} dialogue disentanglement solution. Firstly, we train a model on a multi-participant response selection dataset harvested from the web which is not annotated; we then apply the trained model to perform zero-shot dialogue disentanglement. Without any labeled data, our model can achieve a cluster F1 score of 25. We also fine-tune the model using various amounts of labeled data. Experiments show that with only 10\% of the data, we achieve nearly the same performance of using the full dataset\footnote{Code is released at \url{https://github.com/chijames/zero_shot_dialogue_disentanglement}}.
\end{abstract}

\section{Introduction}
Multi-participant chat platforms such as Messenger and WhatsApp are common on the Internet. While being easy to communicate with others, messages often flood into a single channel, entangling chat history which is poorly organized and difficult to structure. In contrast, Slack provides a thread-opening feature that allows users to manually organize their discussions. It would be ideal if we could design an algorithm to automatically organize an entangled conversation into its constituent threads. This is referred to as the task of~\emph{dialogue disentanglement}~\cite{shen2006thread,elsner2008you,wang2009context,elsner2011disentangling,jiang2018learning,kummerfeld2018large,zhu2020did,li2020dialbert,yu2020online}.

\begin{figure}[!ht]
\includegraphics[width=0.5\textwidth]{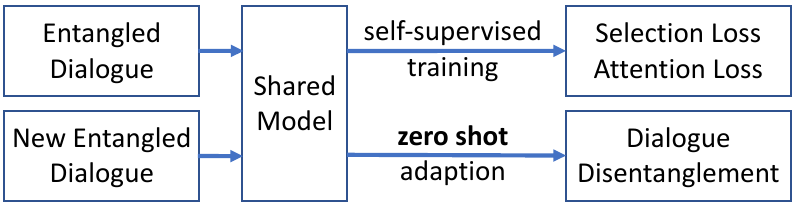}
\caption{This is the high-level flow of our proposed approach.}
\label{fig:flow_chart}
\end{figure}

Training data for the dialogue disentanglement task is difficult to acquire due to the need for manual annotation. Typically, the data is annotated in the~\emph{reply-to} links format, i.e. every utterance is linked to one preceding utterance. The effort is quadratic w.r.t the length of dialogue, partly explaining the sole existence of human-annotated large-scale dataset~\cite{kummerfeld2018large}, which was constructed based on the Ubuntu IRC forum. To circumvent the need for expensive labeled data, we aim to train a self-supervised model first then use the model to perform zero-shot dialogue disentanglement. In other words, our goal is to find a task that can learn implicit~\emph{reply-to} links without labeled data.

\emph{Entangled response selection}~\cite{ws-aaai-dstc20task2} is the task that we will focus on. It is similar to the traditional response selection task, whose goal is to pick the correct next response among candidates, with the difference that its dialogue context consists of multiple topics and participants, leading to a much longer context (avg. 55 utterances).
We hypothesize that:
\begin{quote}
    \emph{A well-performing model of entangled response selection requires recovery of reply-to links to preceding dialogue.}
\end{quote}
This is the only way that a model can pick the correct next response given an entangled context.
Two challenges are ahead of us:
\begin{itemize}
    \item Choosing a design for such a model. Previous work relies on heuristics to filter out utterances to condense context. A model should not rely on heuristics. See \S\ref{sec:entangled_model} and \ref{sec:entangled_attn}.
    \item Even though we can train a well-performing model, how should we reveal the links learned implicitly? See \S\ref{sec:attn_super}.
\end{itemize}

Finally, we want to highlight the high practical value of our proposed method. Consider that we have access to a large and unlabeled corpus of chat (e.g. WhatsApp/Messenger) history. The only cost should be training the proposed entangled response selection model with attention supervision using unlabeled data. The trained model is immediately ready for dialogue disentanglement.
In summary, the contributions of this work are:
\begin{itemize}
    \item Show that complex pruning strategies are not necessary for entangled response selection.
    \item With the proposed objective, the model trained on entangled response selection can perform zero-shot dialogue disentanglement.
    \item By tuning with 10\% of the labeled data, our model achieves comparable performance to that trained using the full dataset.
\end{itemize}

\section{Entangled Response Selection}
\subsection{Task Description}
\label{sec:entangled_desc}
The dataset we use is DSTC8 subtask-2~\cite{ws-aaai-dstc20task2}, which was constructed by crawling the Ubuntu IRC forum.
Concretely, given an entangled dialogue context, the model is expected to pick the next response among 100 candidates.
The average context length is 55 and the number of speakers is 20 with multiple (possibly relevant) topics discussed concurrently. The context is too long to be encoded by transformer-based models~\cite{devlin2018bert,liu2019roberta}. Despite the existence of models capable of handling long context~\cite{yang2019xlnet,zaheer2020big,beltagy2020longformer}, it is difficult to reveal the implicitly learned~\emph{reply-to} links as done in \S\ref{sec:attn_super}.

\begin{table}
  \centering
  \begin{tabular}{@{\extracolsep{0pt}}lcccccc@{}}
    \hline
    {\bf Model}& & R@1 & R@5 & R@10 & MRR\\
    \hline
    Concatenate & & 51.6 & 72.3 & 80.1 & 61.2\\
    \ + aug & & 64.3 & 82.9 & 88.4 & 72.8\\
    Hierarchical & & 50.0 & 72.1 & 81.4 & 60.3\\
    \ + aug & & 65.7 & 84.8 & 91.8 & 74.3\\
    \hline
  \end{tabular}
  \caption{Test set performance of the entangled response selection task. Concatenate is the model with complex pruning heuristics described in \citet{wu2020enhance}.}
  \label{tab:response_selection}
\end{table}

\subsection{Related Work}
To the best of our knowledge, previous works adopt complex heuristics to prune out utterances in the long context~\cite{wu2020enhance,wang2020response,gu2020pre,Dario2020}. 
For example, keeping the utterances whose speaker is the same as or referred to by the candidates. This is problematic for two reasons. 1) The retained context is still noisy as there are multiple speakers present in the candidates. 2) We might accidentally prune out relevant utterances even though they do not share the same speakers. A better solution is to let the model decide which utterances should be retained.

\begin{figure*}[!ht]
\hspace*{-8mm} 
\includegraphics[width=1.1\textwidth]{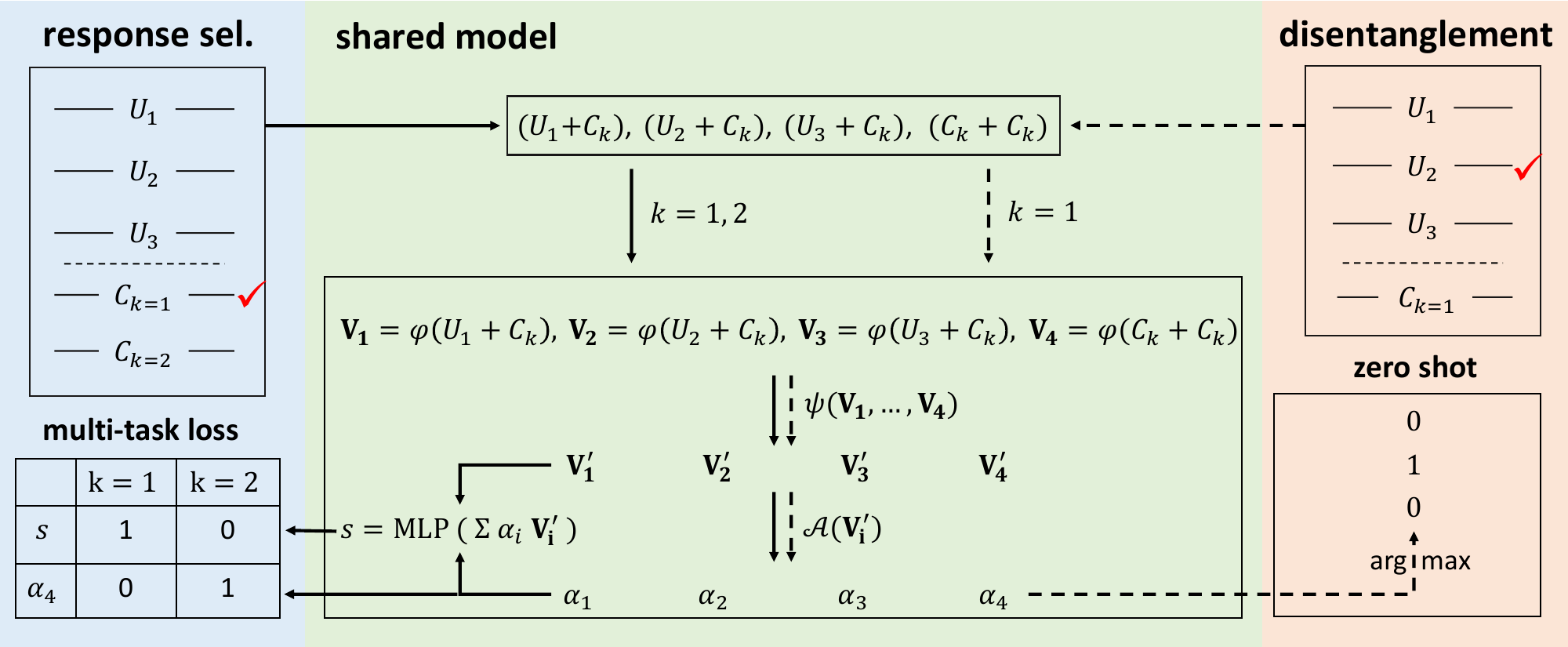}
\caption{\textbf{Solid arrows}:
Given an entangled context $U_{1,2,3}$ and $C_{k=1}$ as the correct next response ($C_{k=2}$ is a negative sample), each pair of the concatenated inputs is encoded separately by $\varphi$ (BERT) to get $\bf V_i$. A context-aware model $\psi$ (transformer) is applied over $\bf V_i$s to generate contextualized $\bf V_i'$. An attention module $\mathcal{A}$ is used to calculate the attention scores $\alpha_i$ and weighted sum $s$. Model is optimized according to the target values for $s$ and $\alpha_4$ in the multi-task loss table. \textbf{Dashed arrows}: Given another entangled context, we know that the current utterance $C_{k=1}$ is replying to $U_2$ by taking the $\arg\max$ of attention scores $\alpha_i$ in a zero-shot manner.}
\label{fig:arch}
\end{figure*}

\subsection{Model (Solid Arrows in Figure~\ref{fig:arch})}
\label{sec:entangled_model}
We use a hierarchical encoder as shown in the middle part of Figure~\ref{fig:arch}.
Suppose the input context is $\{U_i\}_{i=1}^{n}$ and the next response candidate set is $\{C_k\}_{k=1}^{m}$. 
For every candidate utterance $C_k$, we concatenate it with all $U_i$s. For example, we form $n$ pairs for $k=1$, $(U_i+C_1)_{i=1}^{n}$.
Then we use BERT as the encoder ($\varphi$) to encode pairs and get the last layer embedding of the [CLS] token as ${\bf V}_i$:
\begin{gather}
    {\bf V}_i=\varphi(U_i+C_1)_{i=1}^n , \ \  \forall i\in 1\dots n \\
    {\bf V}_{n+1} = \varphi(C_k+C_k)
\end{gather}
While $(C_k+C_k)$ is not necessary for response selection, it is useful later for predicting self-link, which acts as the first utterance of a thread. We will see its role in \S\ref{sec:attn_super}.
Then we use the output embeddings of a one layer transformer ($\psi$) with 8 heads to encode contextualized representations:
\begin{equation}
    \{{{\bf V}_i'\}_{i=1}^{n+1}=\psi(\{{\bf V}_i\}_{i=1}^{n+1}})
\end{equation}
To determine relative importance, we use an attention module ($\mathcal{A}$) to calculate attention scores:
\begin{gather}
    v_i = \textrm{MLP}({\bf V}_i'), \ \  \forall i\in 1\dots n+1 \\
    \{\alpha_i\}_{i=1}^{n+1}=\textrm{softmax}(\{v_i\}_{i=1}^{n+1})
\end{gather}
The final predicted score is:
\begin{equation}
    s = \textrm{MLP}(\sum_{i=1}^{n+1} \alpha_i{\bf V}_i')
\end{equation}
Note that $s$ should be 1 for $C_1$ (the correct next response), and otherwise 0 (row 1 of the multi-task loss table in Figure~\ref{fig:arch}). This can be optimized using the binary cross-entropy loss.

\subsection{Results}
We show the results in Table~\ref{tab:response_selection}. The performance of our approach is comparable to previous work. Note that our model does not use any heuristics to prune out utterances. Instead, the attention scores $\alpha_i$ are decided entirely by the model. 
We also run an experiment using augmented data following~\citet{wu2020enhance}, which is constructed by excerpting partial context from the original context\footnote{For a context of length 50, we can take the first $i\in {1\dots49}$ utterances as a new context and the $i+1$-th utterance acts as the new correct next response. We can sample the negatives randomly from other sessions in the dataset.}. Finally, we want to highlight the importance of the attention module $\mathcal{A}$, where the performance drops by 10 points if removed.

\subsection{Attention Analysis}
\label{sec:entangled_attn}
The empirical success of the hierarchical encoder has an important implication: it is able to link the candidate with one or multiple relevant utterances in the context.
This can be proved by the attention distribution $\alpha_i$.
Intuitively, if $C_k$ is the correct next response (i.e. $k=1$), then the attention distribution should be sharp, which indicates an implicit~\emph{reply-to} that links to one of the previous utterances. In contrast, if $C_k$ is incorrect (i.e. $k\neq 1$), our model is less likely to find an implicit link, and the attention distribution should be flat. Entropy is a good tool to quantify sharpness. Numerically, the entropy is 1.4 (sharp) when $C_k$ is correct and 2.1 (flat) for incorrect ones, validating our suppositions.

Is it possible to reveal these implicit links? The solution is inspired by the labeled data of dialogue disentanglement as elaborated in \S\ref{sec:attn_super}.
 
\begin{table*}
  \centering
    \begin{tabular}{!{\extracolsep{4pt}}lccccccccccc}
    \hline
    \multirow{2}{*}{\bf Setting} & \multirow{2}{*}{$w$}&
    \multirow{2}{*}{$\textrm{data}\%$}&
    \multicolumn{5}{c}{\bf \textsc{cluster}} & \multicolumn{3}{c}{\bf \textsc{link}} \\
    \cline{4-8}
    \cline{9-11}
    & & & VI & ARI & P & R & F1 & P & R & F1\\
    \hline
    \it 1) zero shot
    & \\
    & 0.00 & 0.0 & 62.9 & 14.7 & 2.9 & 0.3 & 0.5 & 41.2 & 39.7 & 40.5 \\
    & 0.25 & 0.0 & 84.4 & 50.1 & 25.9 & 24.8 & 25.3 & 43.7 & 41.4 & 42.2 \\
    & 0.50 & 0.0 & 84.6 & 51.5 & 24.6 & 23.8 & 24.2 & 41.5 & 40.0 & 40.8 \\
    & 0.75 & 0.0 & 84.6 & 49.2 & 23.3 & 23.1 & 23.2 & 41.8 & 40.3 & 41.1 \\
    & 1.00 & 0.0 & 84.3 & 47.5 & 22.9 & 23.0 & 23.0 & 41.6 & 40.1 & 40.9 \\
    \hline
    \it 2) few shot & \\
    \rowcolor[gray]{0.9}[6.5pt][9.6pt]
    finetune & 0.25 & 1 & 89.7 & 60.2 & 26.1 & 33.9	& 29.5 & 65.0	& 62.7 & 63.8 \\
    scratch & - & - & 88.7 & 58.7 & 22.6 & 28.6 & 25.2 & 63.7 & 61.4 & 62.6 \\
    \rowcolor[gray]{0.9}[6.5pt][9.6pt]
    finetune & 0.25 & 10 & 90.6 & 59.8 & 32.4 & 38.4 & 35.1 & 70.5 & 68.0 & 69.3 \\
    scratch & - & - & 90.4 & 61.0 & 32.4 & 36.5 & 34.3 & 70.4 & 67.9 & 69.1 \\
    \rowcolor[gray]{0.9}[6.5pt][9.6pt]
    finetune & 0.25 & 100 & 91.1 & 62.7 & 35.3 & 42.0 & 38.3 & 74.2 & 71.6 & 72.9 \\
    scratch & - & - & 91.2 & 62.1 & 35.6 & 40.3 & 37.8 & 74.0	& 71.3 & 72.6 \\
    \hline
  \end{tabular}
  \caption{$w=0$ indicates pure entangled response selection training. In the few-shot section, scratch is the disentanglement model not trained self-supervisedly on entangled response selection before. The evaluation metrics and labeled data used for fine-tuning are in~\citet{kummerfeld2018large}. Results are the average of three runs.}.
  \label{tab:disentanglement}
\end{table*}

\section{Zero shot Dialogue disentanglement}
\subsection{Task Description}
The dataset used is DSTC8 subtask-4~\cite{kummerfeld2018large}\footnote{It is a disjoint split of the Ubuntu IRC channel as opposed to the one used in \S\ref{sec:entangled_desc}, hence there is no risk of information leak.}.
We want to find the parent utterance in an entangled context to which the current utterance is replying, and repeat this process for every utterance. After all the links are predicted, we run a connected component algorithm over them, where each connected component is one thread.

\subsection{Related Work}
All previous work~\cite{shen2006thread,elsner2008you,wang2009context,elsner2011disentangling,jiang2018learning,kummerfeld2018large,zhu2020did,li2020dialbert,yu2020online} treat the task as a sequence of multiple-choice problems. Each of them consists of a sliding window of $n$ utterances. The task is to link the last utterance to one of the preceding $n-1$ utterances. This model is usually trained in supervised mode using the labeled~\emph{reply-to} links. Our model also follows the same formulation.

\subsection{Model (Dashed Arrows in Figure~\ref{fig:arch})}
We use the trained hierarchical model in \S\ref{sec:entangled_model} without the final MLP layer used for scoring. In addition, we only have one candidate now, which is the last utterance in a dialogue. We use $C_{k=1}$ to represent it for consistency. Note that we only need to calculate $i'=\arg\max_i \alpha_i$. This indicates that $C_{k=1}$ is replying to utterance $U_{i'}$ in the context.

\subsection{Proposed Attention Supervision}
\label{sec:attn_super}
We note that the labeled~\emph{reply-to} links act as supervision to the attention $\alpha_i$: they indicate which $\alpha_i$ should be 1. We call this extrinsic supervision. Recall the implicit attention analysis in \S\ref{sec:entangled_attn}, from which we exploit two kinds of intrinsic supervision:
\begin{compactitem}
    \item If $C_k$ is the correct next response, then $\alpha_{n+1} = 0$ because $C_k$ should be linking to one previous utterance, not itself.
    \item If $C_k$ is incorrect, then it should point to itself, acting like the start utterance of a new thread. Hence, $\alpha_{n+1} = 1$.
\end{compactitem}
We train this intrinsic attention using MSE (row 2 of the multi-task loss table in Figure~\ref{fig:arch}) along with the original response selection loss using a weight $w$ for linear combination $L=(1-w)*L_{res}+w*L_{attn}$. Note that we do not use any labeled disentanglement data in the training process.

\begin{figure}[!ht]
\includegraphics[width=0.5\textwidth]{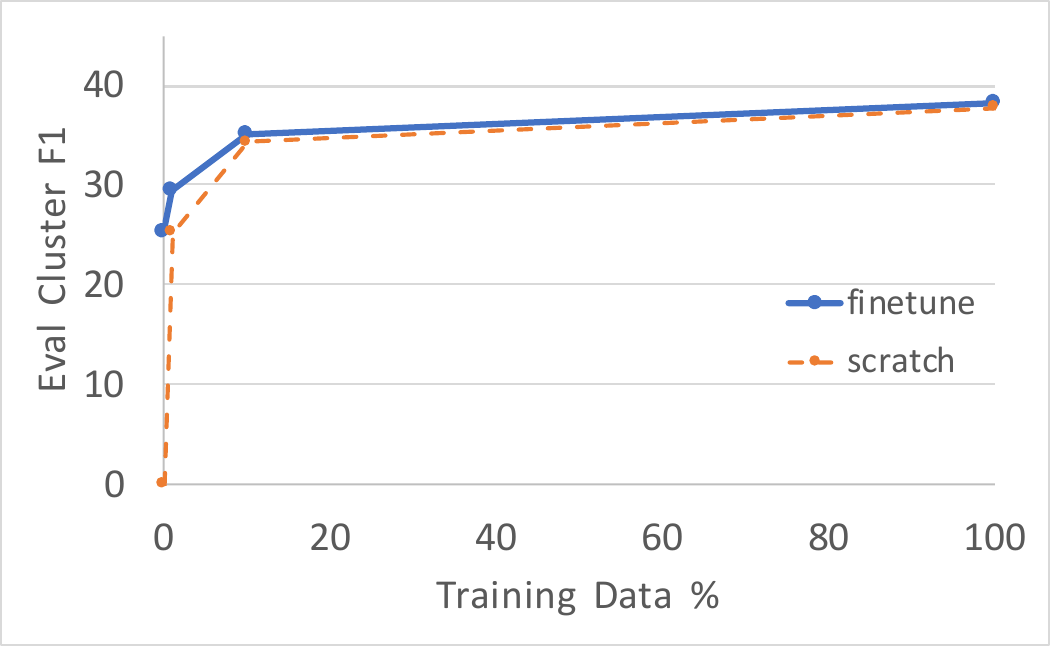}
\caption{Different amounts of labeled data for fine-tuning. The model with self-supervised response selection training outperforms the one trained from scratch.}
\label{fig:trend}
\end{figure}

\subsection{Results}
We present the results in Table~\ref{tab:disentanglement}.
In the first section, we focus on zero-shot performance, where we vary $w$ to see its effect. As we can see, $w=0.25$ gives a close-to-best performance in terms of cluster and link scores. Therefore, we use it for few-shot fine-tuning setting, under which our proposed method outperforms baselines trained from scratch by a large margin. We pick the best checkpoint based on the validation set performance and evaluate it on the test set. This procedure is repeated three times with different random seeds to get the averaged performance reported in Table~\ref{tab:disentanglement}. With 10\% of the data, we can achieve 92\% of the performance trained using full data. The performance gap becomes smaller when more data is used as illustrated in Figure~\ref{fig:trend}.

\subsection{Real-World Application}
Our method only requires one additional MLP layer attached to the architecture of~\citet{li2020dialbert} to train on the entangled response selection task, hence it is trivial to swap the trained model into a production environment. Suppose a dialogue disentanglement system~\cite{li2020dialbert} is already up and running:
\begin{enumerate}
    \item Train a BERT model on the entangled response selection task (\S\ref{sec:entangled_desc}) with attention supervision loss (\S\ref{sec:attn_super}). This is also the multi-task loss depicted in Figure~\ref{fig:arch}.
    \item Copy the weight of the pretrained model into the existing architecture~\cite{li2020dialbert}.
    \item Perform zero-shot dialogue disentanglement (zero-shot section of Table~\ref{tab:disentanglement}) right away, or finetune the model further when more labeled data becomes available (few-shot section of Table~\ref{tab:disentanglement}).
\end{enumerate}
This strategy will be useful especially when we want to bootstrap a system with limited and expensive labeled data.

\section{Conclusion}
In this paper, we first demonstrate that entangled response selection does not require complex heuristics for context pruning. This implies the model might have learned implicit~\emph{reply-to} links useful for dialogue disentanglement. By introducing intrinsic attention supervision to shape the distribution, our proposed method can perform zero-shot dialogue disentanglement. Finally, with only 10\% of the data for tuning, our model can achieve 92\% of the performance of the model trained on full labeled data. Our method is the first attempt to zero-shot dialogue disentanglement, and it can be of high practical value for real-world applications.

\section*{Acknowledgment}
The authors acknowledge the support from Boeing (2019-STU-PA-259).
\bibliography{anthology,custom}
\bibliographystyle{acl_natbib}

\end{document}